\documentclass[letterpaper, 10 pt, conference]{ieeeconf}  

\IEEEoverridecommandlockouts                              

\usepackage[T1]{fontenc}
\usepackage{amsmath}
\usepackage{graphicx}
\usepackage{url}
\usepackage{hyperref}
\hypersetup{colorlinks=true, linkcolor=black, citecolor=black, urlcolor=blue}

\title{\LARGE \bf
SuperEx: Enhancing Indoor Mapping and Exploration \\ using Non-Line-of-Sight Perception
}

\author{Kush Garg$^{1}$ and Akshat Dave$^{2}$
\thanks{$^{1}$Department of Applied Mathematics, Delhi Technological University, 110042, New Delhi, India
        {\tt\small  kushgarg\_23mc079@dtu.ac.in}}%
\thanks{$^{2}$Stony Brook University,
        100 Nicolls Rd, Stony Brook, NY 11794, United States
        {\tt\small dave@cs.stonybrook.edu}}%
}

\begin{document}

\maketitle
\thispagestyle{empty}
\pagestyle{empty}

\begin{abstract}

Efficient exploration and mapping in unknown indoor environments is a fundamental challenge, with high stakes in time-critical settings. In current systems, robot perception remains confined to line-of-sight; occluded regions remain unknown until physically traversed, leading to inefficient exploration when layouts deviate from prior assumptions. In this work, we bring non-line-of-sight (NLOS) sensing to robotic exploration. We leverage single-photon LiDARs, which capture time-of-flight histograms that encode the presence of hidden objects -- allowing robots to look around blind corners. Recent single-photon LiDARs have become practical and portable, enabling deployment beyond controlled lab settings. Prior NLOS works target \emph{3D reconstruction} in static, lab-based scenarios, and initial efforts toward NLOS-aided navigation consider simplified geometries. We introduce \emph{SuperEx}, a framework that integrates NLOS sensing directly into the mapping–exploration loop. SuperEx augments global map prediction with beyond-line-of-sight cues by (i) carving empty NLOS regions from timing histograms and (ii) reconstructing occupied structure via a two-step physics-based and data-driven approach that leverages structural regularities. Evaluations on complex simulated maps and the real-world KTH Floorplan dataset show a 12\% gain in mapping accuracy under $<\!30\%$ coverage and improved exploration efficiency compared to line-of-sight baselines, opening a path to reliable mapping beyond direct visibility. Project webpage: \href{https://super-ex.github.io}{https://super-ex.github.io}


\end{abstract}


\section{Introduction}
Robust exploration and mapping in unknown environments is a fundamental challenge in robotics, with critical applications in indoor search and rescue, medical triage, disaster response, mining, and planetary missions. These scenarios are often time-sensitive, where rapid understanding of the environment is critical. A key limitation of today’s mobile robots is that their vision is limited to what is within their line of sight. Areas occluded by walls, debris, or structural collapse remain unexplored until physically traversed, leaving robots vulnerable to poor decision-making. Existing methods that attempt to predict unseen environments from geometric priors or structural regularities alone often fail catastrophically in the presence of novel layouts. To address this, we leverage non-line-of-sight (NLOS) sensing to enhance indoor mapping and exploration.

When photons emitted by a LiDAR strike a surface, most return directly to the sensor, but a fraction scatter diffusely, bouncing off secondary, hidden surfaces before returning.
Single-photon LiDARs can capture the travel time of each photon returning to the sensor in the form of time-of-flight histograms. Thus, such LiDARs can detect three-bounce signals that encode the presence of otherwise occluded objects -- enabling robots to look around corners. 

A decade of lab research on single-photon time-of-flight sensing has rapidly translated into practical, portable systems. Commodity modules (e.g., ams OSRAM \cite{AMS_TMF882X} and ST Microelectronics \cite{STMicroVL53L4CX}) and widely deployed mobile LiDARs (e.g., Apple iPhone \cite{King2022iPhoneLiDAR}) deliver histogram-based ranging in compact, low-power form factors; and show viability for depth and even NLOS tasks outside controlled labs \cite{AMS_TMF882X,STMicroVL53L4CX,King2022iPhoneLiDAR,Callenberg2021LowCostSPAD}. These trends collectively indicate that single-photon timing histograms are no longer confined to benchtop demonstrations, but are increasingly actionable onboard cues for everyday robots operating in cluttered indoor spaces.

\begin{figure}[t]
  \centering
  \includegraphics[width=\columnwidth]{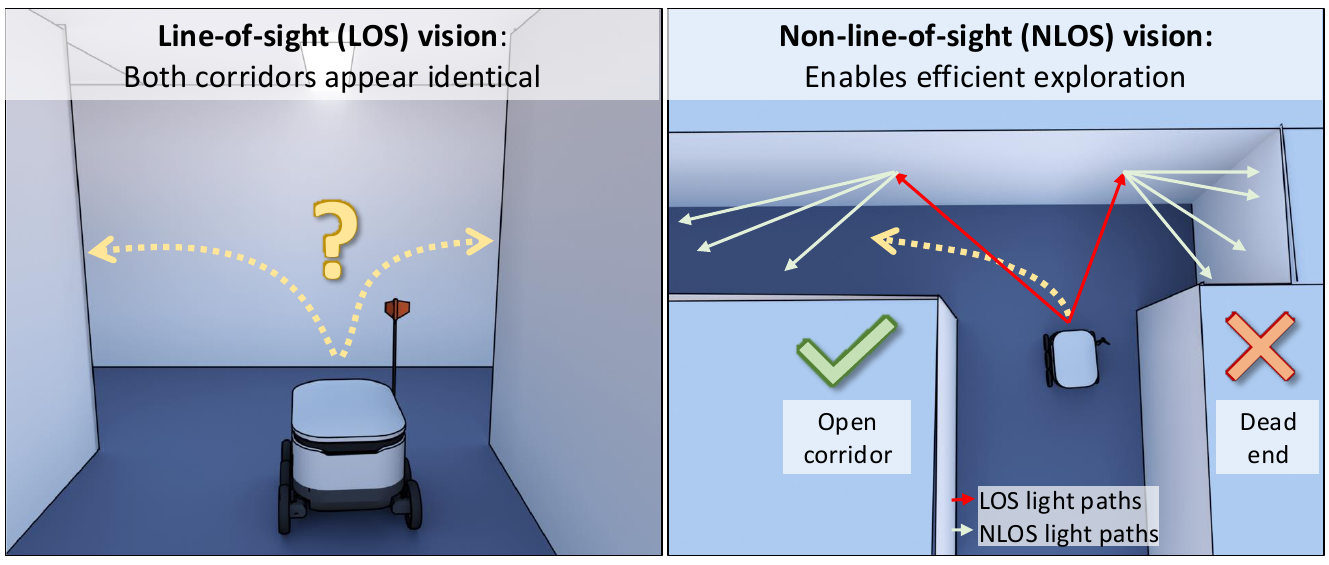}
  \caption{\textbf{Non-line-of-sight (NLOS) perception enables efficient exploration.} Measuring and understanding NLOS light paths enables the robot to distinguish a dead-end from an open corridor, a challenging task to perform from line-of-sight vision alone; and explore the unknown indoor environment efficiently. }
  \label{fig:intro}
\end{figure}

A large body of work uses single-photon LiDAR for non-line-of-sight (NLOS) \emph{3D reconstruction} of hidden objects, typically in laboratory settings with static scenes, planar relay surfaces, and carefully controlled acquisition \cite{Velten2012, OToole2018, Liu2019Phasor}. Young et al. \cite{young2025} explored using NLOS cues for autonomous navigation, but under the simplified geometry of a single L-shaped corner. In contrast, we address NLOS-aided \emph{exploration and mapping} from a moving platform under realistic indoor maps -- by integrating NLOS measurements into the mapping-exploration loop; and leveraging structural regularities through data-driven priors.

To this end, we present SuperEx, a novel framework for NLOS-aided robotic mapping and exploration. We model indoor exploration in the recent framework of probabilistic information gain from global map predictions \cite{ho2025mapex}. Our key insight is to \textit{augment} global map prediction from the captured time-of-flight histograms by 1) carving out empty NLOS regions \cite{Tsai2017} and 2) reconstructing occupied regions. For NLOS reconstruction, we perform physics-based back-projection \cite{Velten2012} followed by data-driven filtering. To evaluate our approach, we perform physics-based emulations of single-photon LiDAR measurements on the real-world KTH Floorplan dataset \cite{ericson2024beyond}. We demonstrate 12\% improvement in mapping accuracy in scenarios with less than 30\% coverage compared to existing line-of-sight only approaches.



\section{Related Work}

\begin{figure}[t]
  \centering
  \includegraphics[width=\linewidth]{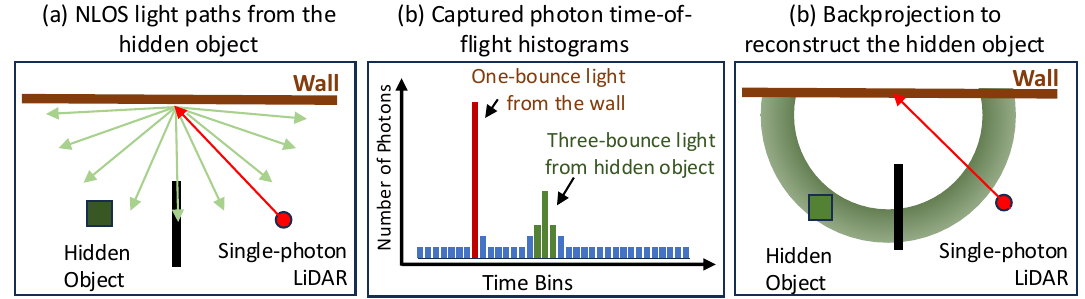}
  \caption{\textbf{Principle of NLOS sensing:} Single-photon LiDAR comprises a pulsed laser, single-photon detector, and timing circuits. (a) When the laser pulse strikes a visible wall, it diffuses, and some of the scattered rays hit the hidden object. Some of the light is scattered back and captured by the sensor as time-of-flight histograms (b), recording the number of photons in each time bin. These measurements are then converted into back-projection maps (c), which represent the likelihood of an object’s presence at a certain distance from the wall.}
  \label{fig:nlosbasics}
\end{figure}

\subsection{Single-photon LiDAR}
Single-photon LiDARs detect individual photon arrivals with picosecond timing, forming per-pixel time-of-flight (ToF) histograms that capture direct and multi-bounce returns, enabling robust 3D at low flux and under ambient light \cite{Lindell2018DeepFusion,Rapp2021HighFlux}. A large body of work leverages SPAD sensors for 3D imaging—using deep sensor fusion and learned reconstructions that exploit full transient signals \cite{Lindell2018DeepFusion,Rapp2021HighFlux} -- and recent efforts broaden the scope with low-cost, miniature cameras for practical shape recovery and downstream perception, including object detection via probabilistic point clouds \cite{Mu2024LowCostSPAD,Goyal2025PPC}. On the hardware side, commodity dToF modules (e.g., STMicroelectronics VL53 series \cite{ST_VL53L7CX}) and widely deployed mobile LiDARs (e.g., iPhone \cite{King2022iPhoneLiDAR}) deliver histogram-based ranging in consumer form factors. Methodologically, researchers exploit raw transient cues beyond peak picking—from “blurred’’ (diffuse-flash) LiDAR for handheld scanning \cite{Behari2025BlurredLiDAR} to planar-deviation detection with tiny sensors \cite{Sifferman2024PlanarDeviation} -- while low-cost SPAD setups demonstrate NLOS tracking and depth \cite{Callenberg2021LowCostSPAD}. Our approach builds on these trends by using histogram measurements not just for standalone depth, but as probabilistic evidence integrated directly into exploration and mapping decisions in complex indoor environments.

\subsection{Non-Line-Of-Sight Imaging}
Non-line-of-sight (NLOS) imaging uses single-photon LiDAR to infer hidden scenes beyond direct view. Seminal results reconstruct “around-the-corner’’ geometry via ultrafast transients \cite{Velten2012} and confocal scanning \cite{OToole2018}, with wave-based methods improving speed, robustness, and sampling generality \cite{Lindell2019,Liu2019Phasor}; in parallel, \emph{data-driven} approaches further advance NLOS reconstruction \cite{Chen2020NLOSEmbeddings,Chopite2020DeepNLOS,Li2023NLOST}. Beyond shape, work demonstrates dynamic tracking \cite{Gariepy2016}, long-range human detection \cite{Chan2017}, and geometric reasoning from first-return photons \cite{Tsai2017}. However, most assume static devices and planar relay surfaces in controlled settings, aiming primarily at 3D imaging; despite progress toward practical hardware—time-gated and array SPAD systems compatible with commodity modules \cite{Buttafava2015,Jin2020}—realistic, cluttered indoor demonstrations remain scarce. Closer to our goal, NLOS cues have been used for navigation under simplified geometry and sensing \cite{young2025}. In contrast, we move beyond “imaging-first’’ pipelines by integrating NLOS evidence into exploration: 360$^\circ$ SPAD transients are back-projected, then filtered/inpainted with data-driven priors, and the resulting beyond-LOS occupancy guides frontier selection, enabling reliable mapping and decisions in complex occluded spaces.
\begin{figure*}[t]
  \centering
  \includegraphics[width=\textwidth]{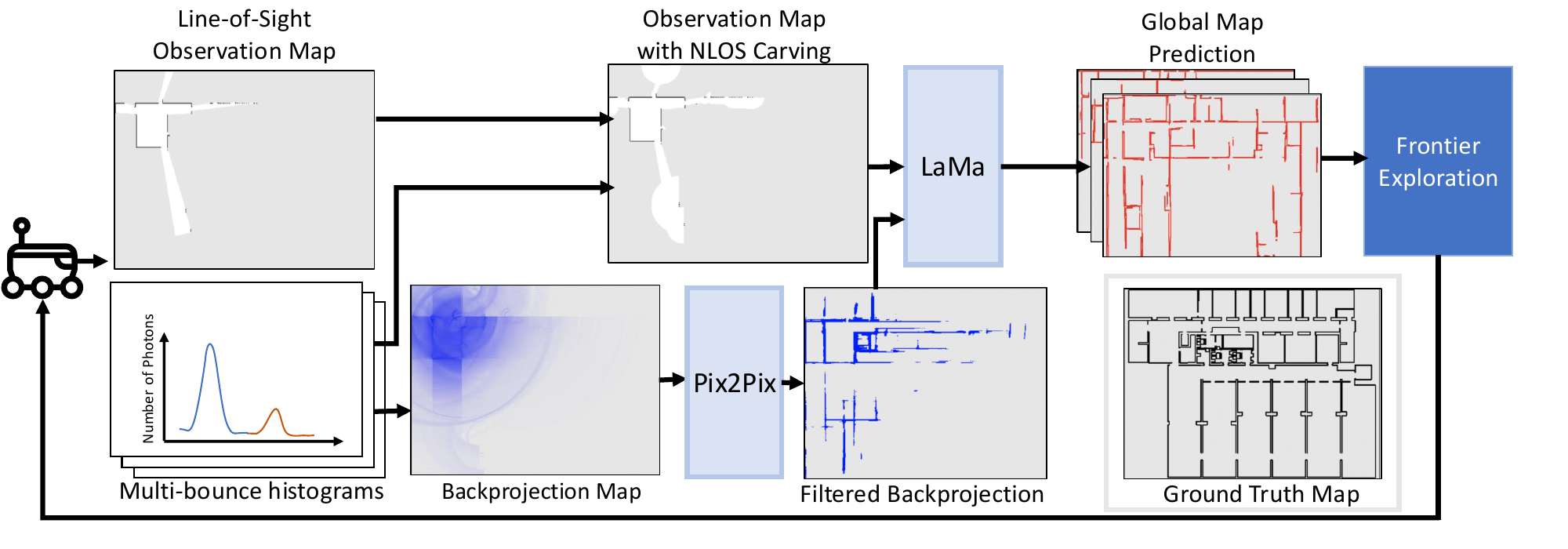} 
  \caption{\textbf{SuperEx pipeline.} The histograms captured by the single-photon LiDAR enable 1) carving out NLOS regions that are empty and 2) backprojection of occupied NLOS, that is filtered with a Pix2Pix network. Both the carved occupancy and filtered backprojection are fed into the Lama network for improved global map prediction, and then for enhanced frontier exploration.}
  \label{fig:pipeline}
\end{figure*}
\subsection{Exploration and Mapping}

The core problem of exploration involves the simultaneous development of environmental understanding and decision-making of where to go next. This decision heavily relies on the accurate knowledge of the environment. The core principle is identifying boundaries between knows and unknown regions called frontiers and selecting the frontier which is expected to give maximum information. Traditional approaches use greedy approach to select the nearest frontiers \cite{Yamauchi1997AFA}. Various variants have been proposed such as viewpoint selection \cite{Desingh2012Viewpoint} and graph-search heuristics \cite{Senarathne2015Incremental}  which model the environment as a graph to optimize exploration globally. Another set of popular approaches use information gain \cite{Stachniss2005Information} , topological \cite{Gomez2019Topological}\cite{Cui2024Frontier}\cite{11007888}, graph based techniques \cite{Ryu2020Graph} and deep learning–based map predictors \cite{Wang2024Improved} coupled with reinforcement learning planners \cite{Leong2023Reinforcement}\cite{Cao2024Deep}. However, these methods are often constrained to short-horizon decisions, limiting their utility in large-scale exploration. To address this, hybrid frameworks like UPEN \cite{Georgakis2022Uncertainty}, IG-Hector \cite{Shrestha2019Learned}, Mapex \cite{ho2025mapex} combine deep learning–based map prediction with classical exploration planning, enabling long-horizon reasoning. UPEN \cite{Georgakis2022Uncertainty} is focused on predicting multiples plausible maps and IG-Hector \cite{Shrestha2019Learned} analyzes amount of area observed from each frontier using a predicted map. MapEx \cite{ho2025mapex} uses LaMa  \cite{suvorov2021resolution} for map prediction and combines both of the ideas using a probabilistic information gain metric as a heuristic for frontier selection. While highly effective, MapEx \cite{ho2025mapex} is limited by its reliance on line-of-sight perception alone. Our work builds directly on it, by introducing NLOS-informed map prediction and exploration. By modifying the LaMa \cite{suvorov2021resolution} inpainting architecture to fuse LOS observations with NLOS reconstructions, we extend the predicted map with higher confidence specially for occluded regions.

\section{NLOS Aided Mapping and Exploration}

We propose a complete pipeline for simulating and integrating non-line-of-sight (NLOS) perception into robotic mapping and exploration. The framework is divided into three modules:
1) Simulation: We develop a physics-based simulator for SPAD-based LiDARs that models multi-bounce photon propagation. This generates transient histograms and corresponding backprojection images that capture indirect reflections in complex environments.
2) Map Reconstruction: We use a sequential pipeline comprising an image-to-image translation model, Pix2Pix \cite{isola2017image}, and an image inpainting model, LaMa \cite{suvorov2021resolution}, to reconstruct NLOS occupancy maps from backprojection images. The reconstructed maps are fused with global map predictions to extend coverage into occluded regions.
3) Mapping and Exploration: We evaluate NLOS-informed mapping within state-of-the-art exploration frameworks. In particular, we adopt the indoor exploration scenarios and configurations introduced in the MapEx benchmark, enabling a direct comparison and demonstrating the benefits of incorporating NLOS perception.

\subsection{Simulation} 

We simulate robotic exploration in an unknown indoor environment, represented as a 2D occupancy grid map. The ground truth occupancy map is denoted as $O_{\mathrm{gt}}$ where each pixel corresponds to a spatial resolution of $0.1 \, \mathrm{m}$. The robot is initialized at a starting pose $o_{\mathrm{start}}$ within the indoor map, without prior knowledge of the environment.  
We simulate a commercially available confocal single-photon avalanche diode (SPAD) LiDAR in 2D. The LiDAR configuration is kept standard, providing full $360^{\circ}$ coverage similar to conventional 2D LiDARs, with an effective range $\mu$. For simplicity, the sensor is modeled as noise-free. The LiDAR emits $n$ rays uniformly in all directions, each ray intersecting the first visible wall at a primary hit point $h_{1}$. The distance of the $i$-th ray to its primary hit point is denoted as $d_{1}^{i}$.  


\begin{figure}[t]
  \centering
  \includegraphics[width=\linewidth]{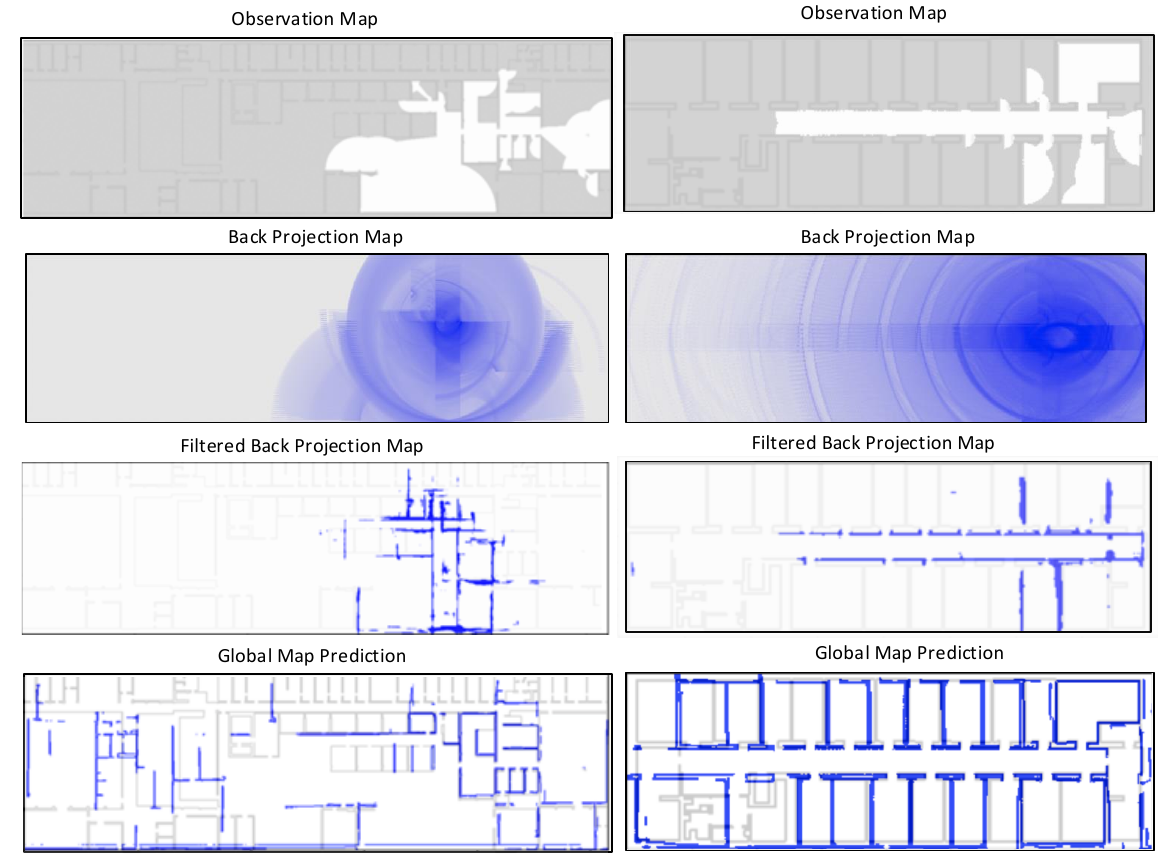}
  \caption{\textbf{Back projection map filtering and global map prediction: } Using the filtered back projection map and the observation map, pipeline predicts a plausible global map.}
  \label{fig:Pix2Pixresult}
\end{figure}

 To simulate diffuse reflection at the wall, we require the surface normal at the hit point. Normals are estimated from the occupancy grid $O_{\mathrm{gt}}$ by performing a 2D convolution with Sobel filters to compute image gradients. The normalized gradient vectors provide an approximation of the wall orientation. To determine the correct outward-facing normal for diffusion, we compute the angle of incidence $\theta_{i}$ between the incoming ray and the candidate normal vectors, and select the one that yields an acute angle. Once the normal $\mathbf{n}_{i}$ at the primary hit point is determined, we diffuse secondary rays in the angular range $\theta \in [-90^{\circ}, 90^{\circ}]$ with respect to $\mathbf{n}_{i}$. These are denoted as $r_{ij}$, where $j$ indexes the secondary ray corresponding to the $i$-th primary ray. Each secondary ray $r_{ij}$ is traced until it intersects another wall at distance $d_{2}^{ij}$.

For each primary ray, we compute a time-of-flight (ToF) histogram $H_{i}(t)$ that encodes the photon arrivals across discrete temporal bins.  The time bin index is defined as  
\begin{equation}
    b = \frac{d}{c \cdot \Delta t},
    \label{eq:bins}
\end{equation}  
where $d$ is the traveled distance by the photon, $c$ is the speed of light, and $\Delta t$ is the time-bin resolution. 

Next, we calculate the photon count for their respective time bins using for multi-bounce returns by combining distance fall-off, angular incidence (cosine factors), surface reflectance $\rho$, and sensor efficiency $\alpha$ following a similar modeling approach as used in Gutierrez et al. \cite{gutierrez2025looking}.




Using this the corresponding number of photons detected  for first bounce is:
\begin{equation}
    N_{i} = \frac{ \alpha\ E_i \, \rho}{\pi} \cdot \frac{\lambda}{h c},
    \label{eq:bounce1}
\end{equation}
where $E_i$ is energy of each ray, $h$ is Planck's constant and $\lambda$ is the wavelength of laser.

Once the primary ray $i$ hits the first surface $h_1^{i}$, light diffuses into $j$ secondary rays, after the 3-bounce process, the number of photons reaching the sensor is: 





\begin{equation}
    N_{2} = \frac{E_i \, \rho^3 \, \cos \theta_{1i} \cos \theta_{2ij} \, (0.1)^2 \, \alpha}{\pi^3 \, d_{1i} \, d_{2ij}^2} \cdot \frac{\lambda}{h c}
    \label{eq:bounce2}
\end{equation}

where $\theta_{1i}$ is angle of incidence of primary ray, $\theta_2^{ij}$ is the angle of incidence of the secondary ray and (0.1) relates to the grid resolution (1 pixel=0.1m). From formula (~\ref{eq:bins}~\ref{eq:bounce1}~\ref{eq:bounce2}) we are able to construct physically realistic histograms.

\subsection{Processing Histograms and Map Reconstruction}

In our LiDAR simulation, the only raw measurements available are time-of-flight (ToF) histograms. In Fig~\ref{fig:nlosbasics}, the first peak in the histogram corresponds to the \emph{line-of-sight (LOS)} return from the primary surface hit, whereas the second peak reflects a \emph{third-bounce} interaction. While the primary hit point can be localized precisely, higher-order peaks encode more ambiguous information: photons arriving in the second peak indicate that an object exists at a certain distance from the first hit point, though its exact position remains ambiguous.

\textit{Space carving.} Here, we leverage \textit{NLOS space carving} developed by Tsai et al.\cite{Tsai2017}. Specifically, for the first filled bin of the second peak, we can assert with certainty that no object lies within that radius of the primary hit point. This insight directly enlarges the robot's observation space, providing free-space guarantees beyond LOS sensing. 

\textit{Filtered Back-projection.} To process the occluded regions further, we perform back-projection \cite{Velten2012}: each histogram bin corresponds to a semicircular uncertainty region with the primary hit point at its center and radius given by the bin index. The back-projection approach accumulates evidence from multiple rays, resulting in a probabilistic NLOS map. While individual back projections along simple walls yield interpretable high-confidence regions, complex indoor structures produce highly cluttered maps that are difficult to interpret directly, as shown in Fig~\ref{fig:Pix2Pixresult}.

To reconstruct meaningful maps, we adopt a sequential pipeline. We employ a Pix2Pix framework \cite{isola2017image} consisting of a U-Net generator and a PatchGAN discriminator. The U-Net architecture enables the retention of high-resolution spatial details during image-to-image translation. For training, we represent the map as a top-down occupancy grid, where the secondary-bounce hit points ($h_2^{ij}$) form the ground-truth supervision. For validation, we train Pix2Pix on the KTH Floorplan dataset (184 floorplans, $\sim$2500 samples), ensuring that training and test sets do not contain floorplans from the same building to promote generalization.

The raw back-projection maps, represented as single-channel float32 tensors normalized to $[-1,1]$, are used as input. Since the wall accounts for less than 1\% of the total image, this class imbalance caused the model to collapse to blank predictions. To mitigate this, we introduce a weighted $\ell_1$ loss emphasizing wall pixels. This results in sharper, more interpretable reconstructions that significantly improve robot confidence in regions beyond LOS regions.

\begin{figure}[t]
  \centering
  \includegraphics[width=\linewidth]{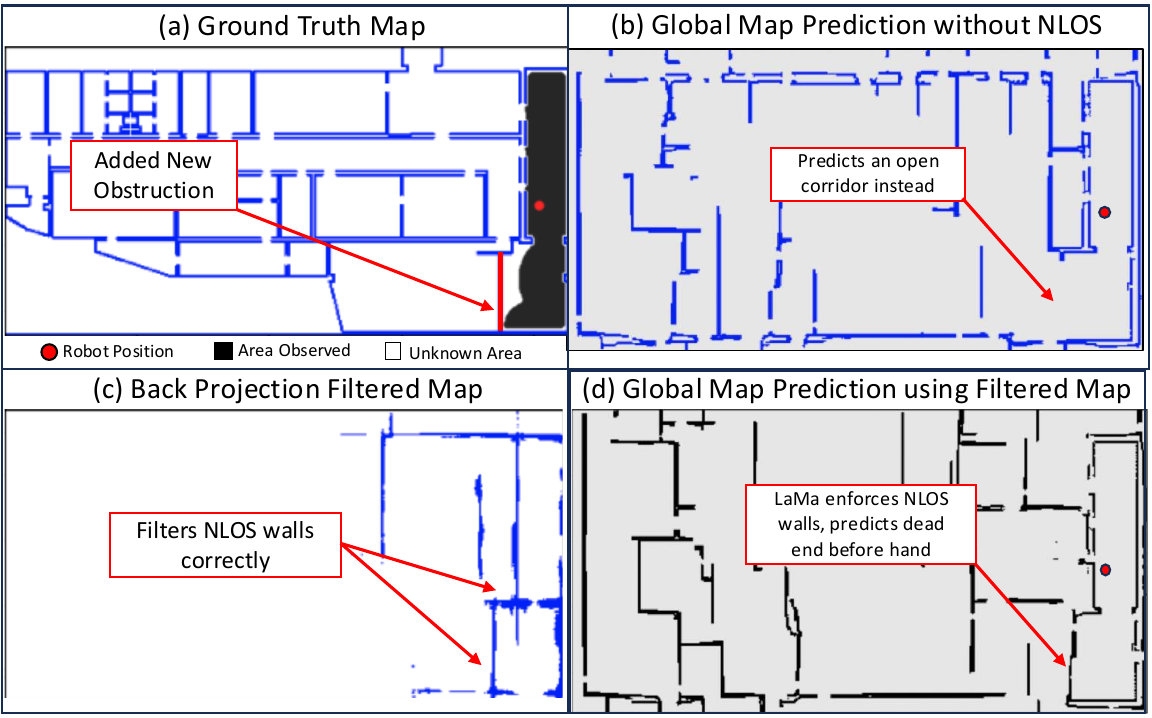}
  \caption{\textbf{Filtered back-projection map during NLOS sensing improves global map prediction over LOS sensing}. (a) As the robot is traversing, an additional obstruction is introduced. (b) With LOS sensing, global map prediction fails to reconstruct the additional obstacle, resulting in an open corridor for exploration. (c) With NLOS sensing, we obtain a filtered back-projection map that correctly reconstructs the additional obstacle. (d) Global map prediction for NLOS takes this filtered back-projection map and correctly predicts the dead end.}
  \label{fig:deadendsim}
\end{figure}

\subsection{Mapping and Exploration}

While LiDAR histograms primarily encode LOS and second-bounce NLOS information, we further enhance reconstruction by integrating \textbf{global map prediction}. Specifically, we leverage the LaMa image inpainting model \cite{suvorov2021resolution}, which combines a large receptive field with strong contextual reasoning, making it well-suited for predicting large unobserved map regions. We augment LaMa’s input by providing (i) the observed occupancy grid, (ii) the Pix2Pix confidence map, and (iii) a binary mask of unobserved areas. The network predicts the complete environment map $O_{pred}$ as shown in ~\ref{fig:Pix2Pixresult}. We train an ensemble of LaMa on diverse dataset to generate multiple plausible map predictions, to estimate uncertainty similar to Mapex \cite{ho2025mapex}. This uncertainty is used to calculate probabilistic information gain metric for each frontier, and the frontier with  maximum probabilistic information gain is selected for traversal. A* \cite{hart1968formal} algorithm is used for path planning to the frontiers, and this sensing-to-planning pipeline is iterated for 1000 exploration steps.





\begin{figure*}[t]
  \centering
  \includegraphics[width=\textwidth]{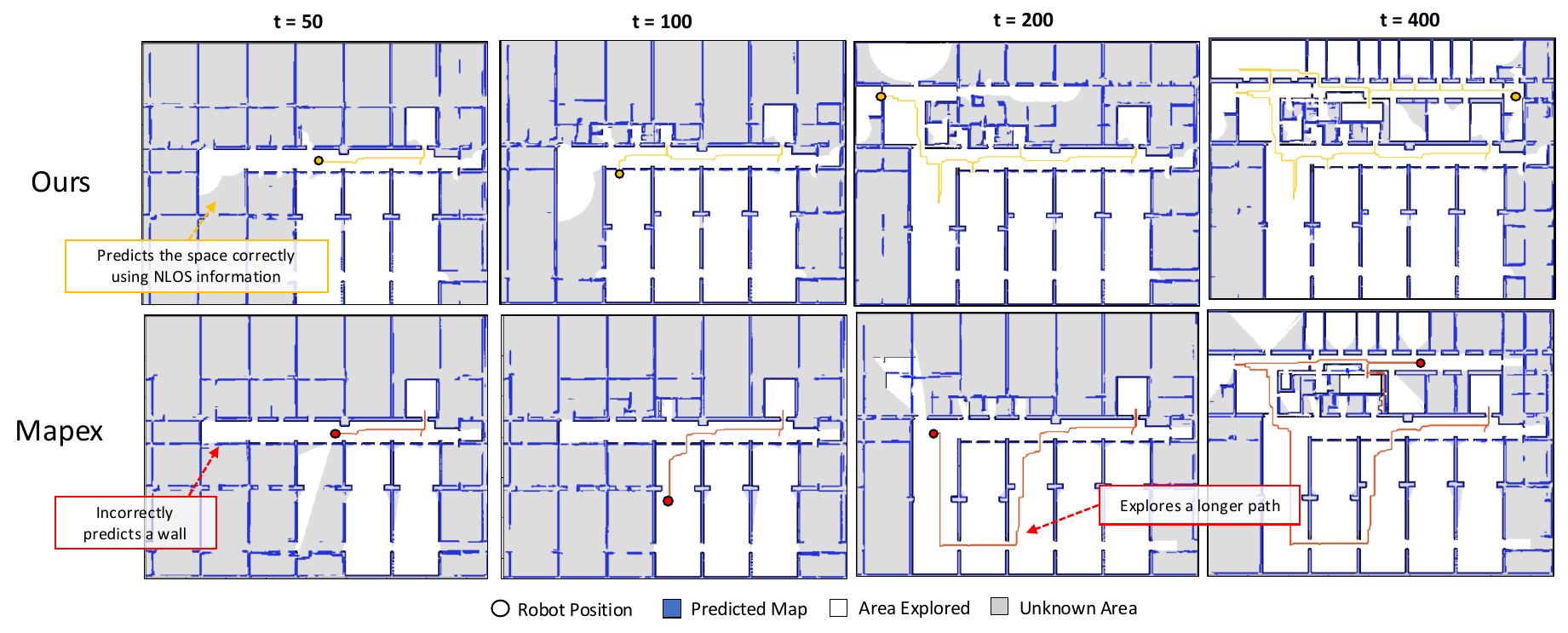} 
  \caption{\textbf{NLOS vs LOS exploration progress over time}: Our approach increases the observation area by seeing NLOS regions, resulting in more coverage and better map prediction. LOS exploration approach MapEx\cite{ho2025mapex} falsely predicts a wall in the global map due to the limited visibility at t=50. As a result, it chooses a suboptimal frontier, resulting in a longer path compared to our NLOS exploration approach.}
  \label{fig:qualitative}
\end{figure*}
\section{Experiments}

\subsection{Implementation Details}
We evaluate our approach on the real-world \textit{KTH Map Dataset} \cite{ericson2024beyond} of indoor structures. We use physics-based simulations, as described in Sec. III.A, to emulate single-photon LiDAR measurements at every robot position from this dataset.  This dataset consists of 184 floor plans collected from 27 different buildings, and is used as $O_{\mathrm{gt}}$ for our experiments.  All methods were tested on 10 randomly chosen maps from the KTH Floorplan Dataset with 4 randomly chosen positions in each map, leading to 40 simulation scenarios over 1000 time steps. The robot is equipped with a SPAD LiDAR with a 20m range and 2500 samples per scan. For the sequential map prediction pipeline, Pix2Pix was trained from scratch on 2,555 images using an 80:20 train-test split. For the LaMa ensemble, the dataset was divided into three subsets, with each LaMa model trained on one subset. We trained LaMa from scratch rather than fine-tuning because we modified the network to accept a 3-channel input (observation map, filtered back-projection map, and binary mask) instead of the standard 4-channel input (RGB image and mask) used in the original LaMa model.

\subsection{Comparisons with line-of-sight exploration and mapping}

We choose 4 baselines for comparison, one is the classical frontier-based approach, and the other three are map prediction-based exploration.
1) Nearest-Frontier: Explores the nearest frontier. 2) IG-Hector: Uses map prediction and information gain metric as a heuristic for exploration. 3) UPEN: Uses an ensemble of map predictions to choose the next frontier. 4) MapEx: Is a combination of both UPEN and IG-Hector that uses an ensemble of map predictions to calculate probabilistic information gain as a heuristic for choosing frontiers. 

We evaluate our approach on the basis of coverage and predicted map accuracy. Coverage is defined as the percentage of area observed by the robot at timestep \emph{t}. Predicted map quality is defined as the Intersection over Union (IoU) of the predicted map and the ground truth map for the percentage coverage of the map. A higher IoU means a more accurate map.

\paragraph{Qualitative Analysis}
Fig~\ref{fig:qualitative}, shows the comparison between our approach and MapEx baseline in exploration at different time steps of simulation. Our approach achieves a larger observation area at each time step using the concept of carving. This additional visibility enables SuperEx to generate more accurate map predictions. For example, at time step 50, SuperEx correctly predicts an empty corridor, whereas MapEx misclassifies it as a dead end due to its line-of-sight constraint, resulting in a longer and less efficient exploration path. Moreover, as the observation map increases, the trajectories of both methods converge. This shows that SuperEx is beneficial in the early stages of exploration when the line of sight information alone is insufficient to predict a reliable map.

Fig~\ref{fig:deadendsim} illustrates the inefficiency of line-of-sight (LOS) methods in environments with added obstructions. When a new wall is introduced beyond the robot’s direct line of sight, the LOS-based prediction fails to detect it. As a result, the robot must traverse the entire L-shaped corridor before recognizing it as a dead end. In contrast, Non-line-of-sight mapping, is able to predict the dead end beforehand, enabling the robot to avoid unnecessary exploration and follow a more efficient trajectory.

\begin{figure}[t]
  \centering
  \includegraphics[width=\linewidth]{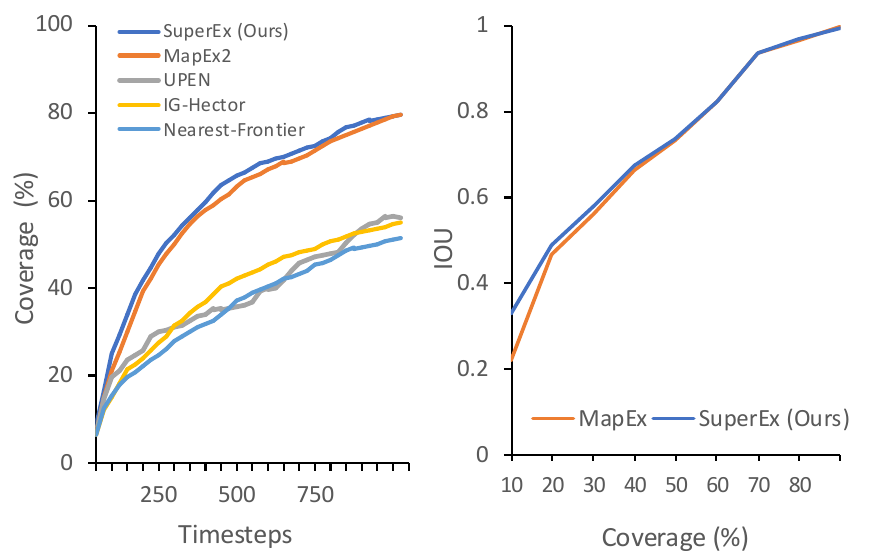} 
  \caption{\textbf{Quantitative evaluation.} SuperEx outperforms LOS exploration baselines in terms of coverage and global map accuracy (IoU).}
  \label{fig:quantitative}
\end{figure}

\paragraph{Quantitative Analysis}
We evaluate SuperEx against baseline methods using Coverage and IoU metrics. SuperEx achieves the best performance overall. As shown in fig~\ref{fig:quantitative}, both SuperEx and MapEx demonstrate significantly higher coverage compared to other baselines. Analysis of the area under the curve for the coverage metric shows that SuperEx outperforms Mapex by 6\% in the initial half of exploration. Beyond this point, coverage gradually converges, as both methods have accumulated sufficient observations to produce reliable map predictions. For IoU as a function of explored area, SuperEx achieves 50\% more accurate map predictions when only 0–10\% of the environment has been observed, and maintains an average IoU improvement of 12\% for cases where up to 30\% of the map has been explored. This shows that NLOS information considerably enhances the initial stages of exploration when line-of-sight observations are insufficient to make reliable predictions.

\subsection{Ablation Studies}
\begin{figure}[t]
  \centering
  \includegraphics[width=\linewidth]{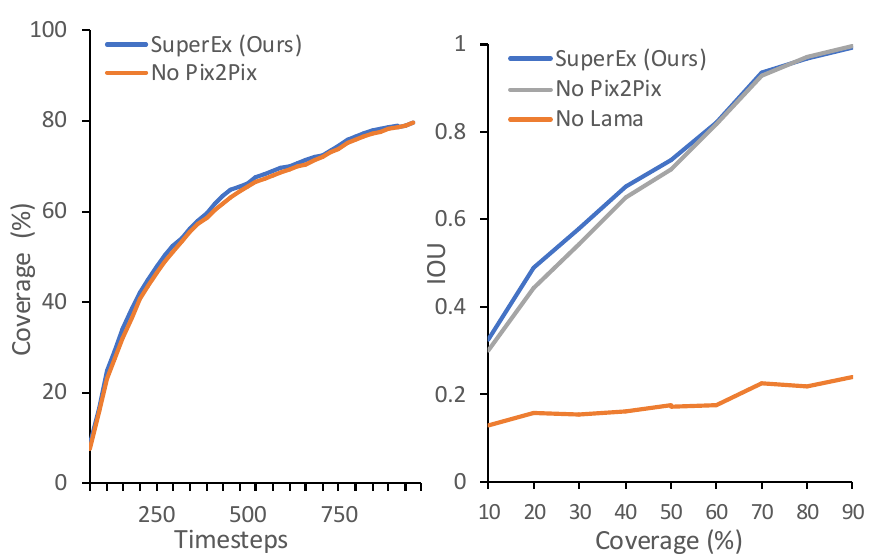} 
  \caption{\textbf{Ablation experiments.} We ablate SuperEx pipeline by 1) removing the learned back-projection map (No Pix2Pix), 2) removing global map prediction (No Lama), and compare the coverage and map prediction metrics, demonstrating the importance of each component.}
  \label{fig:ablation}
\end{figure}

In fig~\ref{fig:ablation} we ablate different components of the sequential map prediction pipeline used by SuperEx to evaluate their individual contributions. We compare our full approach with a variant that excludes back-projection maps but retains carving. Interestingly both have similar coverage metric indicating that carving in itself is a critical component which significantly boosts map prediction for efficient exploration. When using only Pix2Pix with raw back-projection maps as input, it has an average IoU of 0.146, demonstrating its ability to effectively filter the back-projection maps. Furthermore, incorporating Pix2Pix in SuperEx improves IoU, indicating that back-projection maps contribute meaningfully to more accurate map predictions, particularly in the initial stages of exploration.

\begin{figure}[h]
  \centering
  \includegraphics[width=\linewidth]{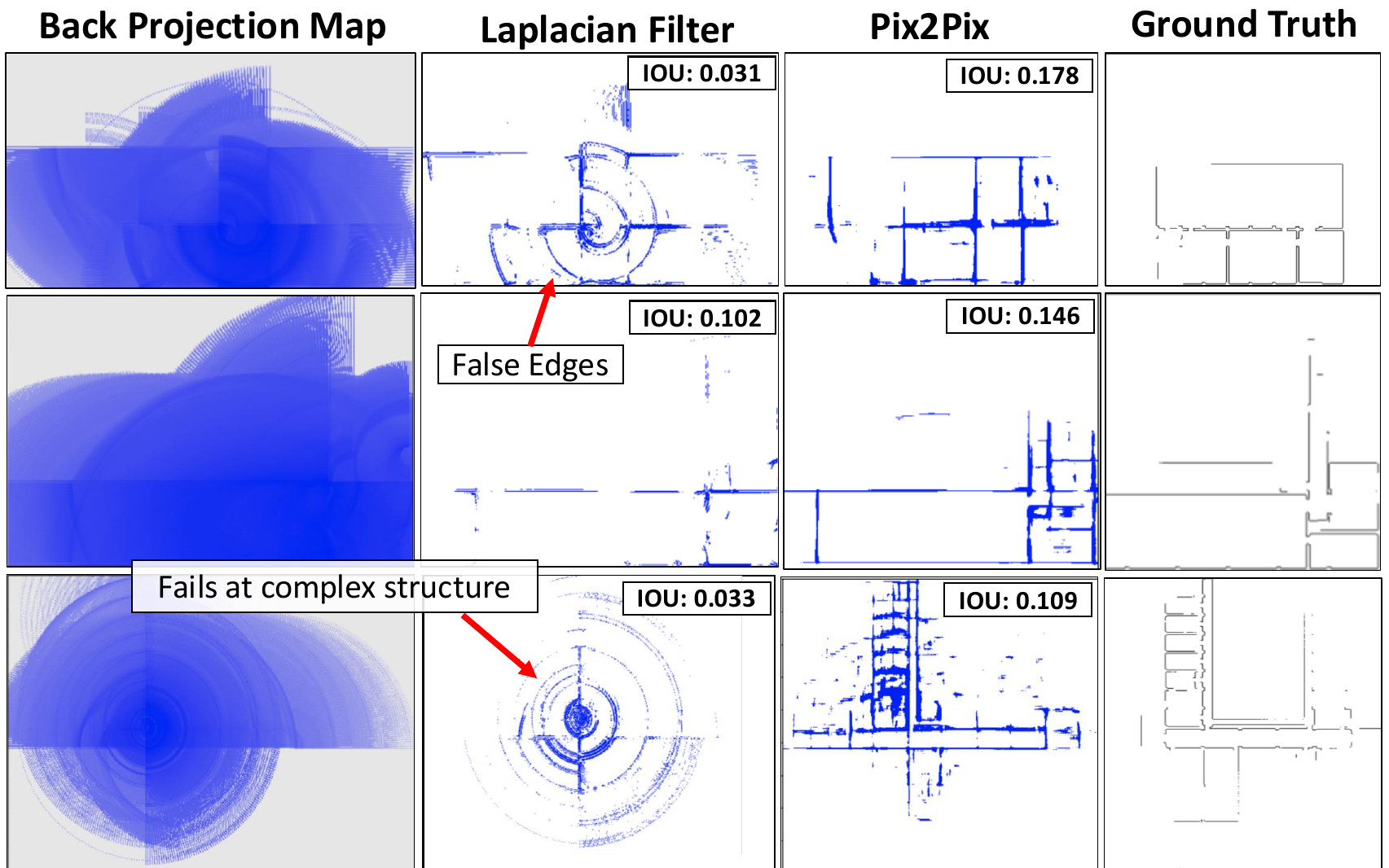}
  \caption{\textbf{Back-projection map filtering comparison between Laplacian filtering \cite{Velten2012} and our module.} We compare our data-driven filtering approach of back-projection maps (termed Pix2Pix) with prior heuristic-based filtering used in NLOS imaging using Laplacian filtering. Evaluated over 400 samples, our Pix2Pix module achieved an average IoU of 0.146, outperforming Laplacian Filter with IoU 0.051.}
  \label{fig:laplace}
\end{figure}

\subsection{Analysis of Back-projection Filtering}

We compare back-projection with Laplacian filtering on the basis of Intersection over Union (IoU) metric. Fig~\ref{fig:laplace} shows that the filter predicts false edges, primarily due reflections from multiple surfaces. With increasing scene complexity, these errors accumulate and significantly degrade the reconstruction quality. Whereas in the case of Pix2Pix \cite{isola2017image}, it is able to filter even in the most complex scenarios with minimal false edges. For quantitative analysis, we computed the IoU of Pix2Pix and Laplacian filtering over 400 back-projection images, , using the Non-Line-of-Sight (NLOS) hit points as ground truth. Pix2Pix achieved an average IoU of 0.146, significantly outperforming the Laplacian filter, which achieved 0.051, demonstrating the effectiveness of our module over Laplacian filtering. 

\section*{Conclusion}
In this work, we propose an end to end pipeline that leverages Non-Line-of-Sight (NLOS) perception enhancing indoor mapping and exploration. The core approach is simulating transient histograms to capture three bounce light transport, form back-projection maps, filter them using data driven deep learning model, and then used for global map prediction. This allows robots to "see around corners" and make more informed exploration decisions. Extensive testing on real world floor plan dataset shows successful filtering of back-projection maps and improved global map prediction and exploration performance. This work paves the way for integrating non-line-of-sight perception into robotic exploration approaches. Future work could explore generalizing our approach to more complex indoor maps through training on larger real-world datasets of 3D indoor maps and through more realistic simulations. Furthermore, we believe that our initial work in this space would enable further research in real-world evaluation by incorporating single-photon LiDARs into robotic platforms

\addtolength{\textheight}{-2cm}   



\bibliographystyle{ieeetr}
\bibliography{references}

\end{document}